\documentclass[conference]{IEEEtran}
\IEEEoverridecommandlockouts
\usepackage{cite}
\usepackage{amsmath,amssymb,amsfonts}
\usepackage{algorithmic}
\usepackage{graphicx}
\usepackage[table]{xcolor}
\usepackage{newtxtext}
\usepackage{newtxmath}
\usepackage{amsmath}
\usepackage{url}
\begin{document}

\title{A Robust Comparison of the KDDCup99 and NSL-KDD IoT Network Intrusion
Detection Datasets Through Various Machine Learning Algorithms}

\author{\IEEEauthorblockN{Suchet Sapre, Pouyan Ahmadi, Khondkar Islam}
\IEEEauthorblockA{\textit{Department of Information Sciences and Technology} \\
\textit{George Mason University}\\
Fairfax, USA \\
ssapre, pahmadi, kislam2@gmu.edu}
}

\maketitle

\begin{abstract}
In recent years, as intrusion attacks on IoT networks have grown exponentially, there is an immediate need for sophisticated intrusion detection systems (IDSs). A vast majority of current IDSs are data-driven, which means that one of the most important aspects of this area of research is the quality of the data acquired from IoT network traffic. Two of the most cited intrusion detection datasets are the KDDCup99 and the NSL-KDD. The main goal of our project was to conduct a robust comparison of both datasets by evaluating the performance of various Machine Learning (ML) classifiers trained on them with a larger set of classification metrics than previous researchers. From our research, we were able to conclude that the NSL-KDD dataset is of a higher quality than the KDDCup99 dataset as the classifiers trained on it were on average 20.18\% less accurate. This is because the classifiers trained on the KDDCup99 dataset exhibited a bias towards the redundancies within it, allowing them to achieve higher accuracies.
\end{abstract}


\section{Introduction}
With the rapid development of the internet, intrusion attacks on IoT networks have been growing exponentially and are a highly pertinent threat in the modern era. Millions of internet users, companies, and national governments are liable to cyberattacks. Consequently, creating sophisticated methods to identify these network intrusions is one of the most prevalent problems in cybersecurity research.  

Network intrusion detection systems (IDSs) heavily rely on data from IoT networks to "learn" patterns that allow them to identify compromised networks. Two prominent datasets used for network intrusion classification are the KDDCup99 and NSL-KDD. The KDDCup99 dataset was created in 1999 by researchers at the University of California, Irvine and was the pioneer intrusion detection dataset. However, it has been empirically shown that the KDDCup99 dataset contains many inefficiencies. The NSL-KDD dataset, which is a subsample of the KDDCup99 dataset, was created by the University of New Brunswick Canadian Institute for Cybersecurity in response to these flaws. 

Throughout this paper, we will refer to two different classification subproblems: binary and type classification. The binary classification refers to identifying whether a network state represents an intrusion or not (no intrusion / normal = 0, intrusion = 1). The type classification refers to identifying whether a network state represents one of five different intrusion types: \textit{Normal} (no intrusions), \textit{DoS}, \textit{Probe}, \textit{R2L}, and \textit{U2R}. These intrusion types are defined in the "Datasets (IV)" section. In the past, there have been very few exhaustive comparisons of both the KDDCup99 and NSL-KDD datasets. As a result, this study hopes to shed light on the similarities and differences between these datasets in regards to how various Machine Learning (ML) classifiers train and perform on them. We used four ML classifiers for this study: the Na\"ive Bayes Classifier (NBC), Support Vector Machines (SVMs), Random Forests, and Artificial Neural Networks (ANNs).

\section{Related Research}
Artificial Intelligence (AI) and ML have been readily used to analyze and classify data in both the KDDCup99 and NSL-KDD network intrusion detection datasets. The most common methods of analysis in the past have been classical ML algorithms such as SVMs and NBCs, however, recently, the focus has shifted to the use of ANNs. For instance, a team of researchers was able to obtain around an 80\% accuracy for the binary classification of the NSL-KDD dataset with an ANN. This result is approximately the same as the result we obtained using an ANN \cite{NSLKDD1}. 

In regard to comparing the quality of both datasets, another group analyzed ensemble ML model's ability to classify IoT intrusion attacks in both of these datasets. They found that in general, the classifiers trained on the KDDCup99 dataset obtained a higher accuracy than those trained on the NSL-KDD dataset. However, this team utilized feature selection when preprocessing their data, which demonstrates why they were able to obtain characteristically higher accuracies of 99.42\% and 98.70\% on the KDDCup99 and NSL-KDD datasets respectively for the binary classification task \cite{Comparison1}.

Finally, one of the most relevant research studies to the contents of this paper was conducted by the original creators of the NSL-KDD dataset. This team constructed the NSL-KDD dataset as a means of improving the KDDCup99 dataset. Namely, they removed redundant and repetitive records from the KDDCup99 dataset in hopes of improving its quality. This group also tested numerous ML algorithms on their newly created NSL-KDD dataset and observed lower accuracies on it in comparison to the KDDCup99 dataset for the binary classification task. They attributed this lower accuracy to the increased rigor of the NSL-KDD dataset as represented by the removal of "easy-to-classify" records \cite{Comparison2}. 

\section{System Model}
Figure 1 shows the system model for this study starting from the dataset and ending with the evaluation of the various ML classifiers.
\begin{figure}[h]
\centering
\includegraphics[width=0.45\textwidth]{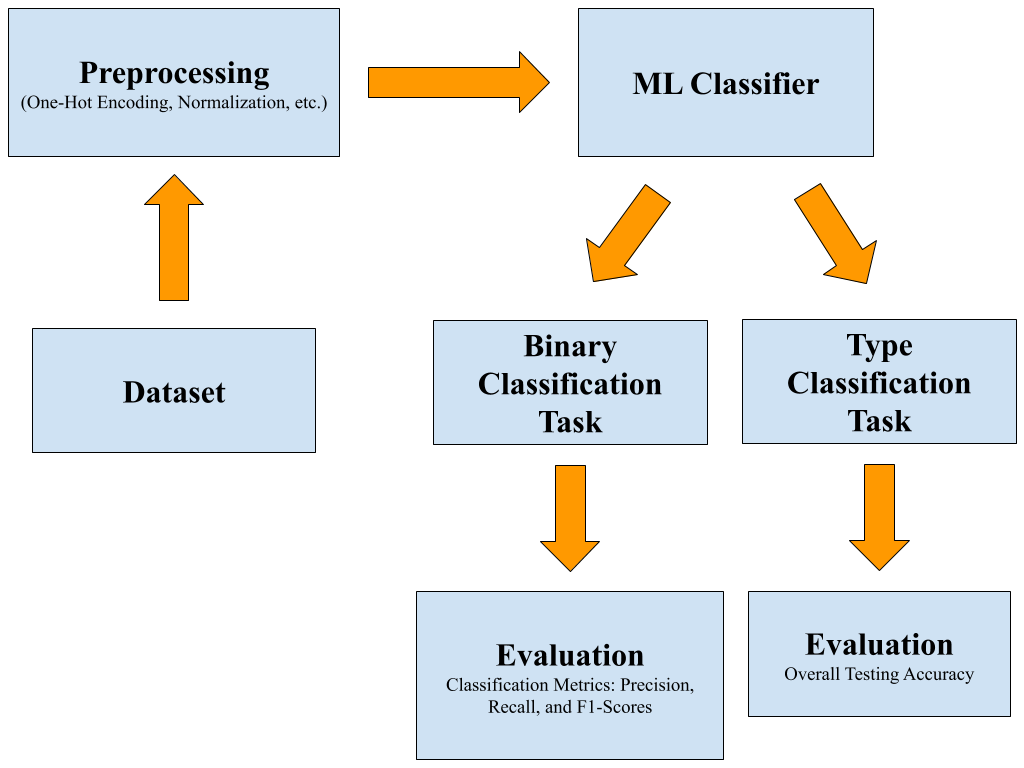}
\caption{Diagram of System Model}
\centering
\end{figure}
\section{Datasets}
\subsection{Overview}
Both the KDDCup99 and NSL-KDD datasets contain five main intrusion types or categories: Normal (No intrusion), DoS, Probe, U2R, and R2L. These intrusion types are defined below,
\begin{enumerate}
    \item \textbf{Normal}: Networks that do not contain any intrusions.
    \item \textbf{DoS}: Shutting down a network by flooding it with information and requests.
    \item \textbf{R2L}: Unauthorized access from a remote machine
    \item \textbf{U2R}: Intruder attempts to gain access to a normal user account
    \item \textbf{Probe}: Surveillance intrusions on networks \cite{IntrusionAttacks}.
\end{enumerate}
In addition, we will also consider the binary categorization for this dataset (i.e. 0 representing when a network intrusion does NOT occur and 1 representing when a network intrusion DOES occur). These two categorization types will be referenced throughout the paper as "type categorization" and "binary categorization," respectively.
\vspace{12pt}
\subsubsection{KDDCup99}
The KDDCup99 is the original IoT network intrusion dataset that was created in 1999. The motivation behind the dataset's creation was to improve the capability of IDSs. Namely, with this dataset, cybersecurity researchers could train algorithms to better identify when a network intrusion occurs based on quantitative and categorical data regarding the state of the network itself. With respect to the training and test datasets, we used the pre-sampled 10\% subset of the KDDCup99 training dataset \textit{("kddcup.data\_10\_percent.gz")} as our experimental training dataset and the full KDDCup99 test dataset \textit{("corrected.gz")} as our experimental test dataset. This choice was made in order to reduce the computational power required to process and handle the full training dataset. Henceforth, whenever we say "KDDCup99 training dataset", we are referring to the 10\% subset of the full KDDCup99 training dataset \cite{IntrusionAttacks}. 

One of the most important aspects of this dataset is the imbalanced class distribution. Notably, the extremely low number of both R2L and U2R intrusion types present within it. Figure 2 and Table 1 display the class distribution within this dataset, illustrating the relatively low number of R2L and U2R attack types in comparison to the other three. Table 2 displays the class distribution in the context of the binary categorization. It is evident that the number of intruded networks far outnumbers the number of normal networks. 

\begin{table}[htbp]
\caption{Intrusion-Type Class Distribution in KDDCup99 Training Dataset}
\renewcommand{\arraystretch}{1.45}
\vspace{-5mm}
\begin{center}
\begin{tabular}{|c|c|c|c|c|c|}
\hline
    & Normal & DoS & R2L & U2R & Probe \\
\hline
Frequency & 97,278   & 391,458  & 1,126  & 52 & 4,107 \\ 
\hline
\end{tabular}
\end{center}
\end{table}

\begin{figure}[h]
\centering
\includegraphics[width=0.4\textwidth]{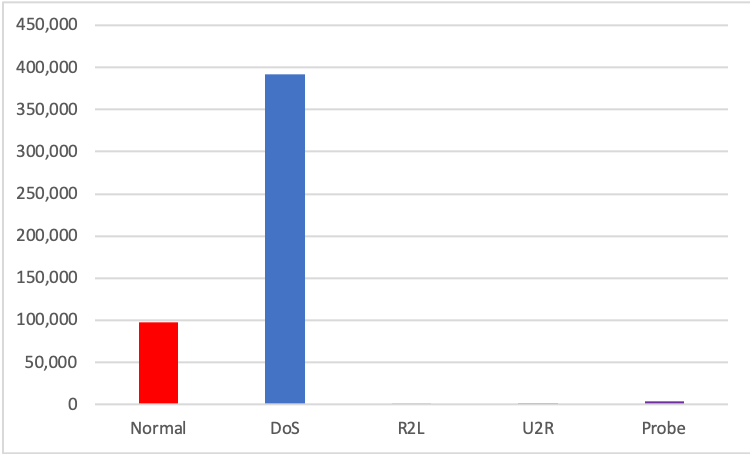}
\caption{Intrusion-Type Class Distribution in KDDCup99 Training Dataset}
\centering
\end{figure}

\begin{table}[htbp]
\caption{Binary Class Distribution in KDDCup99 Training Dataset}
\renewcommand{\arraystretch}{1.45}
\vspace{-5mm}
\begin{center}
\begin{tabular}{|c|c|c|}
\hline
    & Normal & Intrusion \\
\hline
Frequency & 97,278 & 396,743 \\
\hline
\end{tabular}
\end{center}
\end{table}

\subsubsection{NSL-KDD}
The NSL-KDD dataset is the improved version of the KDDCup99 dataset. In essence, researchers in the University of New Brunswick Canadian Institute for Cybersecurity removed redundant records from the original KDDCup99 dataset to create a more concise and efficient intrusion detection dataset. Concerning the training and test datasets, we used the full NSL-KDD training \textit{("KDDTrain+.TXT")} and test \textit{("KDDTest+.TXT")} datasets as our experimental training and test datasets. We were able to do this because the NSL-KDD dataset is much smaller in comparison to the full KDDCup99 dataset \cite{NSLKDDDataset}.

Although there is still a class imbalance between intrusion types, namely, the low number of U2R and R2L intrusions, the fact that there are far fewer instances of the other three intrusion types mitigates the effect of this class imbalance in comparison to the KDDCup99 training dataset. Figure 3 and Table 3 illustrate this. Moreover, Table 4 displays the class distribution in the context of the binary categorization. It is important to note that there is a much greater class balance between normal and intruded networks in comparison to the KDDCup99 training dataset. 

\begin{table}[htbp]
\caption{Intrusion-Type Class Distribution in NSL-KDD Training Dataset}
\renewcommand{\arraystretch}{1.45}
\vspace{-5mm}
\begin{center}
\begin{tabular}{|c|c|c|c|c|c|}
\hline
    & Normal & DoS & R2L & U2R & Probe \\
\hline
Frequency & 67,343 & 45,927 & 995 & 52 & 11,656 \\ 
\hline
\end{tabular}
\end{center}
\end{table}

\begin{figure}[h]
\centering
\includegraphics[width=0.45\textwidth]{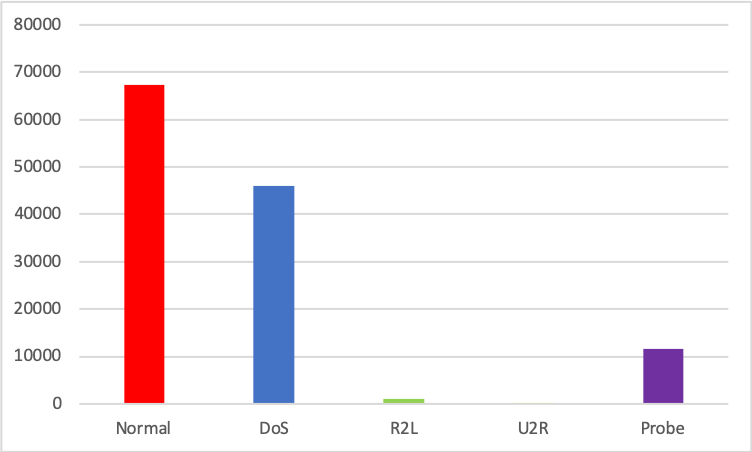}
\caption{Intrusion-Type Class Distribution in NSL-KDD Training Dataset}
\centering
\end{figure}

\begin{table}[htbp]
\caption{Binary Class Distribution in NSL-KDD Training Dataset}
\renewcommand{\arraystretch}{1.45}
\vspace{-5mm}
\begin{center}
\begin{tabular}{|c|c|c|}
\hline
    & Normal & Intrusion \\
\hline
Frequency & 67,343 & 58,630 \\ 
\hline
\end{tabular}
\end{center}
\end{table}
\subsection{Preprocessing}
Because both datasets contained categorical features, we one-hot encoded all of these categorical columns so that the ML algorithms could "understand" the data.

The train and test datasets for both the KDDCup99 and NSL-KDD datasets were normalized to values between 0 and 1 by the L2 or Euclidian normalization. When the L2 normalization is applied on vector $\vec{x}$, such that $\vec{y} = \frac{\vec{x}}{\|\vec{x} \|}$, the sum of the squares of the components of $\vec{y}$ (the normalized vector) equals 1. The L2 normalization was used instead of the L1 normalization because it has been demonstrated that the former has many advantages that improve the performance of ML classifiers. 


\section{Machine Learning Methods}
This section describes the theory and implementation of the four ML algorithms that we used in this study: the Na\"{i}ve Bayes Classifier (NBC), Support Vector Machines (SVMs), Random Forests, and Artificial Neural Networks (ANNs). As a note, all of the ML classifiers were trained on normalized data.
\subsection{Na\"{i}ve Bayes Classifier}
\subsubsection{Overview}
The NBC is a probabilistic classifier that utilizes the Bayes theorem. Its naivety comes from the rather strong assumptions of independence among the features in a particular dataset. Bayes theorem can be expressed as the following mathematical relationship:
\begin{equation}
P(C_k|\vec{x}) = \frac{P(\vec{x}|C_k)P(C_k)}{P(\vec{x})},
\end{equation}
where $\vec{x}$ = ($x_1$, $x_2$, ... , $x_n$) is a data point consisting of $n$ independent features, $C_k$ represents class $k$, and $P(C_k|\vec{x})$ is the probability that data point $\vec{x}$ belongs to class $C_k$. 
The NBC algorithm applies Bayes theorem across all of the features within the dataset. In this manner, the class $\hat{y}$ that data point $\vec{x}$ belongs to (as predicted by the NBC) can be expressed as
\begin{equation}
  \hat{y} = \underset{k \in \{1,...,K\}}{\text{argmax}} P(C_k) \prod_{i=1}^{n} P(x_i | C_k).
\end{equation}
Because some of the features within both datasets were continuous values, we used the Gaussian NBC which simply assumes that these continuous features are distributed according to the Gaussian distribution. Additionally, for the multi-class portion of this classification task, we used the Multinomial NBC.

Going into this study, the NBC served as our standard of comparison as it is known to be less accurate than the other three classifiers used in this study \cite{NBC}.
\vspace{12pt}
\subsubsection{Implementation}
As previously mentioned the Gaussian NBC was used for both the binary and type classification tasks. Additionally, for the type classification task, the Multinomial variant of the Gaussian NBC was used. 

\subsection{Support Vector Machines}
\subsubsection{Overview}
SVMs are ML algorithms that attempt to use hyperplane-based vectors in order to separate the data`s labels with maximal margin. With this hyperplane separation, the SVM is then able to "understand" the spatial location of the classes within the dataset \cite{SVM}.
\vspace{12pt}
\subsubsection{Implementation}
Because SVMs are traditionally used for binary classifications, we only used the SVM in order to classify whether or not a particular network contained an intrusion. This means we did not use the SVM to classify the intrusion type. This is reflected in the results section. 

SVMs are known to be computationally expensive. As a result, we only used a 0.05\% random sample of both the KDDCup99 and NSL-KDD training datasets to train our SVM. Furthermore, we utilized the linear kernel for our SVM.

\subsection{Random Forest}
\subsubsection{Overview}
Random Forests or random decision trees function by creating numerous "mini" decision trees or estimators and repeatedly training these estimators on a particular dataset. The Random Forest evaluates output by essentially taking the majority vote of these estimators. More formally, classification occurs by taking the average (or the majority vote) of the estimators for a given input sample $x'$ as represented by the mathematical function:

\begin{equation}
    \hat{f} = \frac{1}{B} \sum_{b=1}^{B} f_{b}(x`)
\end{equation}

Here, $\hat{f}$ represents the class that sample $x`$ belongs to (as predicted by the Random Forest), $B$ is the number of estimators, and $f_b$ is a function representing what class the particular estimator $b$ predicts sample $x'$ to be in \cite{RandomForest}. 
\vspace{12pt}
\subsubsection{Implementation}
We used 5 estimators in our implementation of the Random Forest algorithm. Furthermore, because Random Forests are also computationally expensive, we only trained it on a 0.1\% random sample of both the KDDCup99 and NSL-KDD training datasets. The Random Forest algorithm was adaptable to both the binary and type classification tasks. Figures 4 and 5 show sample shallow estimators from our KDDCup99 and NSL-KDD Random Forests, respectively.

\begin{figure}[h]
\centering
\includegraphics[width=0.30\textwidth]{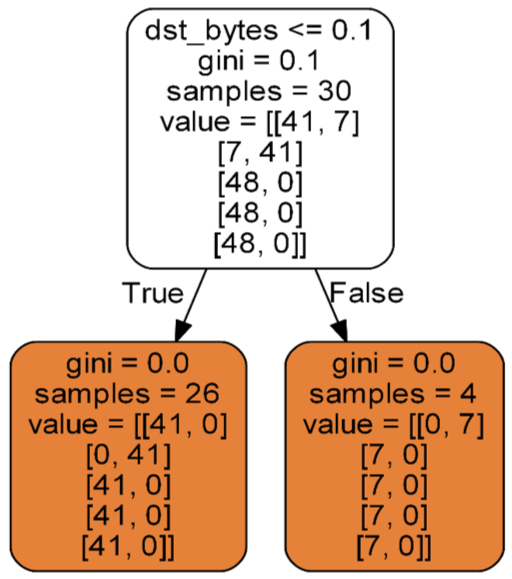}
\caption{Random Forest Estimator - KDDCup99}
\centering
\end{figure}

\begin{figure}[h]
\centering
\includegraphics[width=0.30\textwidth]{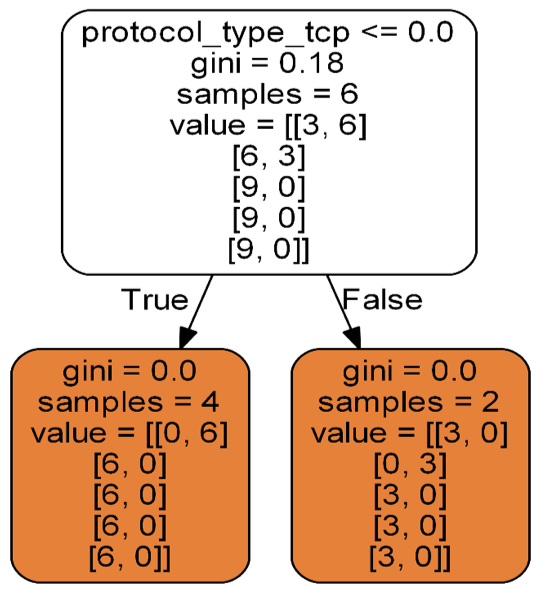}
\caption{Random Forest Estimator - NSL-KDD}
\centering
\end{figure}

\subsection{Artificial Neural Network}
\subsubsection{Overview}
ANNs mimic the processing techniques of the human brain. They do this by passing "information" through a series of weighted neuron layers (nodes) which are fine-tuned until the ANN can successfully replicate the patterns and relationships within datasets' features and labels. This is similar to how our brain gradually adjusts neural connections to learn new skills.

Looking at the mathematical nature of a single node, this connectivity is clear. If there are four nodes each containing the input scalars $x_1, x_2, x_3, x_4$, respectively, that all pass information into one node, there will be four connections. Each of these connections will have a particular "weight", $w_1, w_2, w_3, w_4$, respectively. To replicate the human brain, the one neuron that is receiving the information from the other four has to scale each of the inputs ($x_i$'s) by the weight of their connection. Then, the ANN sums all of these weighted inputs adds a scalar bias and applies an activation function which helps introduce non-linearity to the model. In all, this can be represented mathematically as:
\begin{equation}
    \text{Neuron Output} = f(b + \sum\limits_{i} w_ix_i).
\end{equation}
Here, $f$ is the activation function (sigmoid, ReLU, etc.) and $b$ is the scalar bias. 
\vspace{12pt}
\subsubsection{Implementation}
The topology of our ANN consisted of two hidden layers of size 100 nodes and one output layer of size 5 or 1 nodes depending on whether the classification was type or binary, respectively. Figures 6 and 7 depict the topology of the binary and type classification ANN, respectively. Both hidden layers were activated by the ReLU function defined as,

\begin{equation}
    \text{ReLU}(x) = \text{max}(0, x), 
\end{equation}
The 1 node output layer was activated by the sigmoid function defined as,
\begin{equation}
    \text{sigmoid}(x) = \frac{1}{1+e^{-x}},
\end{equation}
and finally, the 5 node output layer was activated by the softmax function defined as,
\begin{equation}
    \begin{split}
    \text{SoftMax}(z)_i = \frac{e^{z_i}}{\sum_{j=1}^{N} e^{z_j}} \: \text{for} \: i = 1, ... , N \: \\  \text{and} \: z = (z_i, ... , z_N) \in \mathbb{R}^N.
    \end{split}
\end{equation}
This function can take a vector $z$ of dimension $N$ with real values and normalize that vector such that the sum of its components is 1. After normalization, the components of $z$ are probabilities representing the extent to which the classifier believes a particular data point belongs to each class \cite{Softmax}.

\begin{figure}[h!]
\centering
\includegraphics[width=0.42\textwidth]{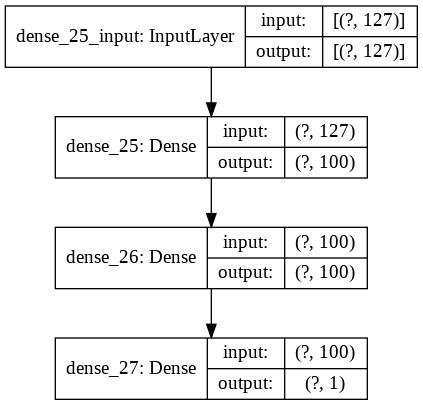}
\caption{Binary Classification ANN Topology}
\centering
\end{figure}

\begin{figure}[h!]
\centering
\includegraphics[width=0.42\textwidth]{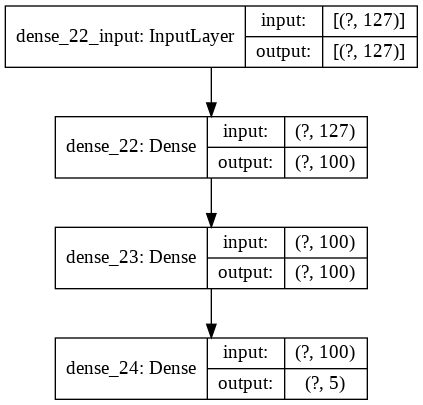}
\caption{Type Classification ANN Topology}
\centering
\end{figure}

Two primary training specifications were present in the ANN's we used in this study: validation-loss based patience and model checkpoints. \textbf{Validation-loss based patience} refers to how many epochs we let our ANN train for. Specifically, we allowed our ANN to keep training until the validation-loss did not decrease for a chosen number of epochs, in our case, we chose this number to be 20 (patience = 20). The \textbf{model checkpoints} specification allowed us to save the best model from the ANN's entire training time based on the validation-loss. To clarify, consider an ANN that trained for 200 epochs, if the validation-loss was the best at epoch 58, the ANN from epoch 58 would be saved and used for the results. 

The Adam Optimizer was chosen here as it has been proven to work for a vast majority of ML classification tasks \cite{Adam}. Additionally, we employed the Categorical Cross Entropy (CCE) loss function defined below:

\begin{equation}
    \text{CCE Loss} = -\sum_{c=1}^{M} y_{o,c} \log(p_{o,c}).
\end{equation}
Here, $M$ is the number of different classes, $y$ is the binary indicator (0 or 1) that the class label $c$ is correct for observation $o$, and $p$ is the probability, calculated by the classifier, that observation $o$ belongs to class $c$ \cite{Loss}. Figures 8 and 9 depict sample training/testing accuracy and loss curves in the training procedure for both the KDDCup99 and NSL-KDD dataset-based classifiers during the binary classification task. It is important to note that the KDDCup99-ANN's validation loss function decreases over an interval while the NSL-KDD-ANN's validation loss function is nearly always increasing. These graphs are cut off at 20 epochs.

\begin{figure}[h]
\centering
\includegraphics[width=0.47\textwidth]{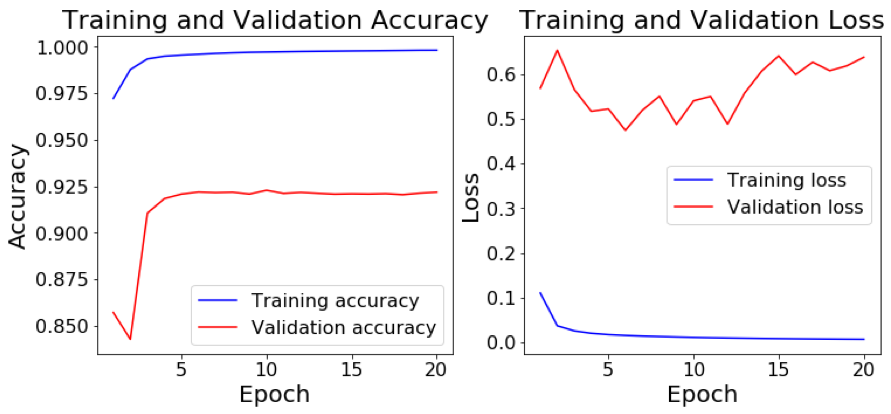}
\caption{Traning/Testing Loss \& Accuracy Curve - KDDCup99}
\centering
\end{figure}

\begin{figure}[h]
\centering
\includegraphics[width=0.47\textwidth]{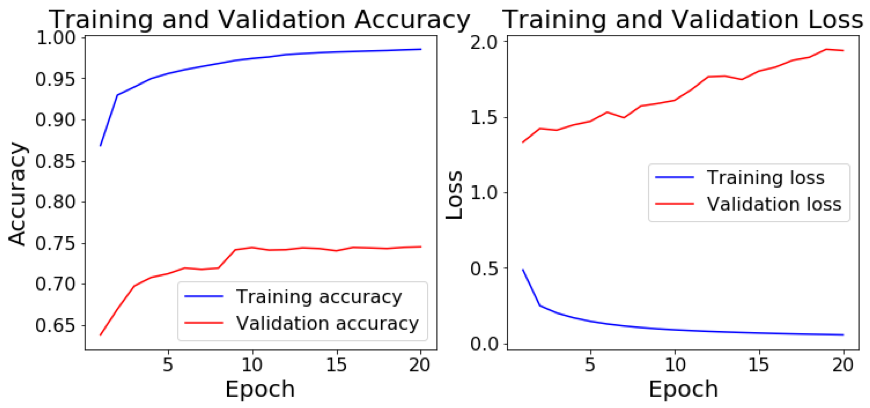}
\caption{Traning/Testing Loss \& Accuracy Curve - NSL-KDD}
\centering
\end{figure}
\section{Results}
\subsection{Evaluation Criteria}
Two separate trials were conducted to evaluate the performance of the ML classifiers. \textbf{Firstly}, we had to evaluate their performance in the type classification task. To do this we utilized an overall testing accuracy (the classifiers' accuracy on the testing portions of the KDDCup99 and NSL-KDD datasets) which is displayed in tables 5 and 6. This was done because generally, classification metrics such as precision, recall, and f1-scores are used to evaluate binary classification tasks. To complement these tables, we also displayed ANN's results in confusion matrices (Figures 10 and 11). \textbf{Secondly}, we had to evaluate the ML classifiers' performance in the binary classification task. To do this, we used the aforementioned classification metrics: precision, recall, and f1-scores. These results are displayed in tables 7 and 8. Each of these metrics is based upon the number of predicted false positives, false negatives, true positives, and true negatives defined as follows,
\begin{enumerate}
    \item False positive ($FP$): A sample that is predicted as positive but is actually negative; 
    \item False negative ($FN$): A sample that is predicted as negative but is actually positive;
    \item True positive ($TP$): A sample that is predicted as positive and is actually positive;
    \item True negative ($TN$): A sample that is predicted as negative and is actually negative.
\end{enumerate}
Here positive refers to a network that contains an intrusion while negative refers to a network that does not. Then we can define the classification metrics as,
\begin{equation}
    \text{Precision}\: (P) = \frac{TP}{TP+FP} 
\end{equation}
\begin{equation}
    \text{Recall}\: (R) = \frac{TP}{TP+FN}
\end{equation}
\begin{equation}
     \text{F1-Score} = 2 \cdot \frac{P \cdot R}{P+R}.
\end{equation}

Because the training and testing datasets for both the KDDCup99 and NSL-KDD datasets were separated by the creators, a cross-validation was unnecessary.

\subsection{Tabular Results}
\vspace{-5mm}
\begin{table}[h!]
\caption{Type Classification Accuracies for KDDCup99}
\renewcommand{\arraystretch}{1.45}
\vspace{-5mm}
\begin{center}
\begin{tabular}{|c|c|c|c|c|c|}
\hline
    & ANN & SVM & NBC & Random Forest & Average \\
\hline
    Accuracy & \textbf{92.39\%} & \cellcolor{gray} & 89.94\% & 92.21\% & 91.51\% \\ 
\hline

\end{tabular}
\end{center}
\end{table}
\begin{figure}[h!]
\centering
\includegraphics[width=0.45\textwidth]{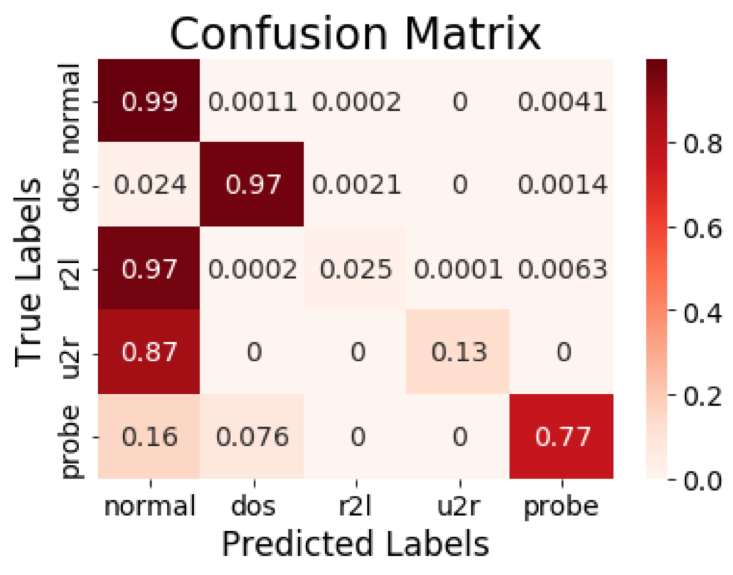}
\caption{Confusion Matrix for ANN (Type Classification) - KDDCup99}
\centering
\end{figure}

\begin{table}[h!]
\caption{Type Classification Accuraies for NSL-KDD}
\renewcommand{\arraystretch}{1.45}
\vspace{-5mm}
\begin{center}
\begin{tabular}{|c|c|c|c|c|c|}
\hline
    & ANN & SVM & NBC & Random Forest & Average \\
\hline
Accuracy & \textbf{78.51\%} & \cellcolor{gray} & 61.08\% & 74.41\% & 71.33\% \\ 
\hline
\end{tabular}
\end{center}
\end{table}

\begin{figure}[h!]
\centering
\includegraphics[width=0.45\textwidth]{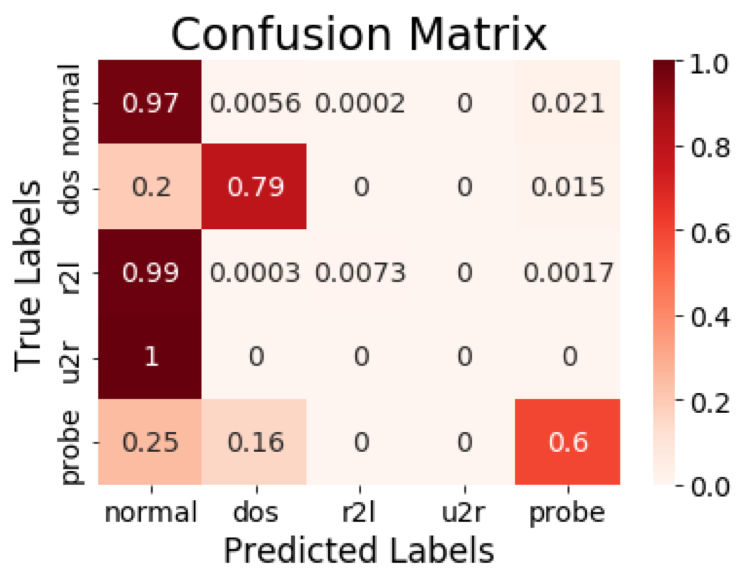}
\caption{Confusion Matrix for ANN (Type Classification) - NSL-KDD}
\centering
\end{figure}

\begin{table}[h!]
\caption{Binary Classification  Metrics for KDDCup99}
\renewcommand{\arraystretch}{1.45}
\vspace{-5mm}
\begin{center}
\begin{tabular}{|c|c|c|c|c|c|}
\hline
    & ANN & SVM & NBC & Random Forest & Average \\
\hline
Precision & 0.9985 & 0.9833 & 0.9937 & \textbf{0.9987} & 0.9936 \\ 
\hline
Recall & 0.9112 & \textbf{0.9339} & 0.8537 & 0.9084 & 0.9018 \\ 
\hline
F1-Score & 0.9529 & \textbf{0.9579} & 0.9185 & 0.9514 & 0.9452 \\ 
\hline
\end{tabular}
\end{center}
\end{table}

\begin{table}[h!]
\caption{Binary Classification Metrics for NSL-KDD}
\renewcommand{\arraystretch}{1.45}
\vspace{-5mm}
\begin{center}
\begin{tabular}{|c|c|c|c|c|c|}
\hline
    & ANN & SVM & NBC  & Random Forest & Average \\
\hline
Precision & 0.9661 & 0.8839 & 0.9672 & \textbf{0.9683} & 0.9464  \\ 
\hline
Recall & 0.6205 & \textbf{0.8142} & 0.1746 & 0.6158 & 0.5563\\ 
\hline
F1-Score & 0.7557 & \textbf{0.8476} & 0.2957 & 0.7528 & 0.6630  \\ 
\hline
\end{tabular}
\end{center}
\end{table}
\section{Discussion}
\subsection{Type Classification Accuracies}
With respect to the type classification testing accuracies (displayed in Tables 5 and 6), it is clear that the ANNs performed the best across both datasets. This was expected as ANNs have been shown to have a better ability to detect patterns within datasets in comparison to the other ML algorithms used in this study (NBCs, SVMs, and Random Forests).  More surprising however, is the difference in type classification accuracy across the classifiers trained by the KDDCup99 and NSL-KDD datasets. On average, the classifiers trained on the KDDCup99 dataset displayed a 20.18\% greater accuracy than those trained on the NSL-KDD dataset. This accuracy difference can be attributed to the fact that the KDDCup99 contains many redundant records which are extremely easy for classifiers to identify. 

\subsection{U2R and R2L Identification}
Figures 10 and 11 display the confusion matrices for the ANN's classification results. One unusual aspect about these confusion matrices is the difference in the classification accuracy for R2L and U2R attacks. Specifically, the ANN trained on the KDDCup99 dataset correctly predicted 2.5\% of R2L and 13\% of U2R attacks while the ANN trained on the NSL-KDD dataset only predicted 0.73\% of R2L and 0\% of U2R attacks. This result goes against one of the primary intentions of the creation of the NSL-KDD dataset. Namely, the increased class balance in the NSL-KDD dataset should have made detecting these infrequent intrusion types easier, however, our results show that it might, in fact, be more difficult. This suggests that some of the records that were removed from the KDDCup99 dataset to create the NSL-KDD dataset may have been essential to properly distinguishing U2R and R2L attacks from the three other intrusion types. Notably, the ANN trained on the NSL-KDD dataset primarily classified R2L and U2R attacks as Normal networks. This shows that there may be underlying problems in the NSL-KDD dataset that is causing this seemingly irrational judgement. 

\subsection{Binary Classification Accuracies}
The binary classification task sought to identify whether a particular network state was normal or was representative of an intrusion attack. Surprisingly, across both the KDDCup99 and NSL-KDD dataset-trained classifiers the SVMs performed the best with f1-scores of 0.9579 and 0.8476, respectively. This result was surprising because SVMs are not as sophisticated as Random Forests or ANNs. Looking specifically at precision and recall values, it is evident that the Random Forests had the best precision values while the SVMs had the best recall values. The ANNs and Random Forests did not perform the best in any of the three categories (precision, recall, or f1-score).

More generally, the classifiers trained on the NSL-KDD dataset had an average f1-score of 0.6630 while the classifiers trained on the KDDCup99 dataset had a much higher average f1-score of 0.9452. A deeper look at this difference reveals that the primary setback for the NSL-KDD-trained classifiers were their low recall values. Namely, the average KDDCup99 classifier recall value was 0.9018 while the average NSL-KDD classifier recall value was much lower at 0.5563. This means that the NSL-KDD classifiers tended to predict many more false negatives. Overall, however, the fact that the NSL-KDD classifiers had a lower performance indicates that the NSL-KDD dataset contains a much more rigorous set of data, which is essential to the strengthening of IDSs. 

\subsection{Principle Component Analysis}
\begin{figure}[h]
\centering
\includegraphics[width=0.45\textwidth]{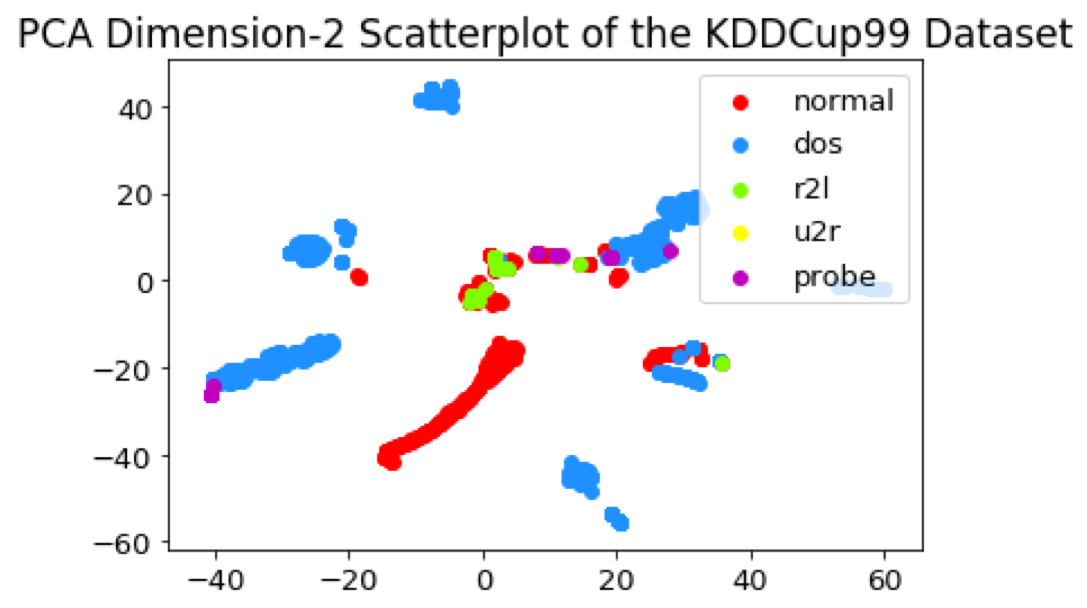}
\caption{PCA Visualization of KDDCup99 Dataset}
\centering
\end{figure}

\begin{figure}[h]
\centering
\includegraphics[width=0.45\textwidth]{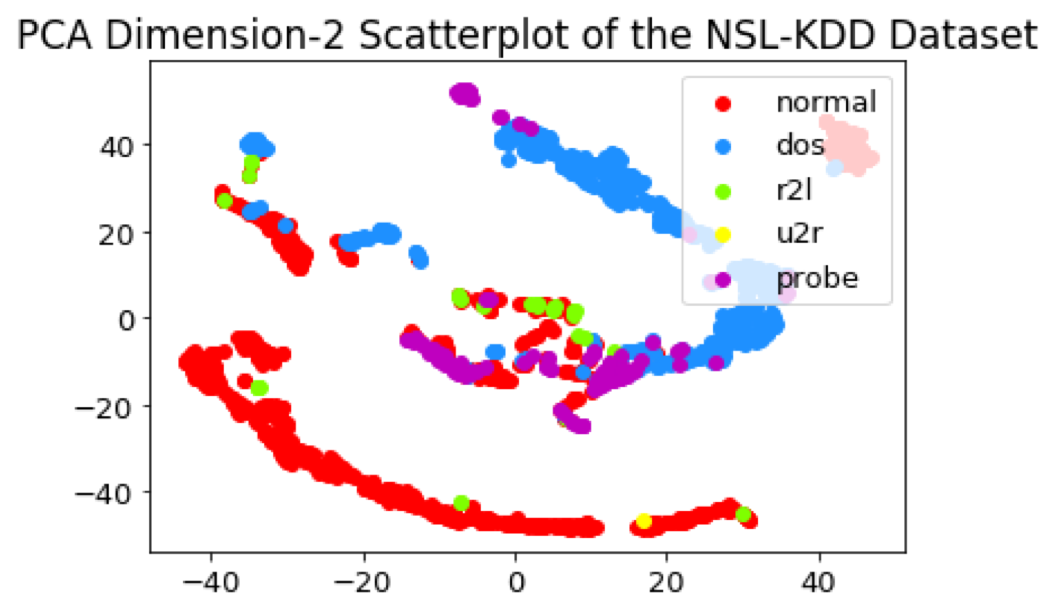}
\caption{PCA Visualization of NSL-KDD Dataset}
\centering
\end{figure}

Figures 12 and 13 display the Principle Component Analysis plots of a random 0.4\% subset of the KDDCup99 dataset and a random 1\% subset of the NSL-KDD dataset, respectively. The Principle Component Analysis allowed us to reduce the dimensionality of the datasets so that they could be visualized on graphs. We observed that the KDDCup99 dataset contained a greater degree of separation between the different intrusion types while the NSL-KDD dataset contained much more overlap. This is consistent with our results as data that has a higher degree of separation between classes is easier to classify. However, this increased spatial separation present in the KDDCup99 dataset is probably indicative of redundant records that are geometrically "overlapping". 

\section{Conclusions \& Future Research}
The quality of the NSL-KDD dataset is significantly higher than that of the KDDCup99 dataset. This conclusion is based off the metrics used to evaluate the performance of the various ML classifiers. 
\vspace{12pt}
\begin{enumerate}
    \item \textbf{Type classification accuracies}: The respective testing type classification accuracies of the ML classifiers trained on the NSL-KDD dataset were much lower than the classifiers trained on the KDDCup99 dataset. This demonstrates how the redundant records in the KDDCup99 dataset enables algorithms to perform with a higher accuracy.
    \item \textbf{Binary classification metrics} (Precision, Recall, and F1-scores): It was observed that the average f1-score for the ML classifiers trained on the KDDCup99 dataset, 0.9452, was much higher than the F1-score for the ML classifiers trained on the NSL-KDD dataset, 0.6630. This further indicates that the ML classifiers perform better on the KDDCup99 dataset due to the redundant and "easy-to-classify" records within it.
\end{enumerate}

\vspace{12pt}

One interesting aspect of our results was the relatively low classification accuracy of the R2L and U2R intrusion types by the classifiers trained on the NSL-KDD dataset. Typically, researchers have found that the NSL-KDD dataset allows for a higher classification accuracy for these intrusion types due to the decreased class imbalance, however, our results showed the opposite. This is likely due to differences in data sampling or classifier topology/architecture.

In the future, our team plans to develop "stacked" or ensemble ML classifiers that are specifically geared towards having a higher accuracy on the NSL-KDD dataset as it is of a significantly higher quality than the KDDCup99 dataset. Specifically, we hope to improve the characteristically low recall values our classifiers obtained on the NSL-KDD dataset which means being able to mitigate the number of false negatives. Finally, we hope to expand our comparative analysis to a wider set of ML classification algorithms for greater completeness.

\section*{Acknowledgment}

We would like to acknowledge the UCI Knowledge Discovery in Databases Archive for providing us with the KDDCup99 dataset and the University of New Brunswick Canadian Institute for Cybersecurity for providing us with the NSL-KDD dataset.

\end{document}